\documentclass[letterpaper, 10 pt, conference]{ieeeconf}  

\IEEEoverridecommandlockouts                              

\usepackage{colortbl}
\usepackage{xcolor}
\usepackage[utf8]{inputenc} 
\usepackage[T1]{fontenc}    
\usepackage{hyperref}       
\usepackage{url}            
\usepackage{booktabs}       
\usepackage{amsfonts}       
\usepackage{nicefrac}       
\usepackage{microtype}      
\usepackage{xcolor}         
\usepackage{graphicx}
\usepackage{multirow}
\usepackage{bm}
\usepackage{pifont}
\usepackage{subfigure}
\usepackage{amsmath}
\usepackage{multirow}
\usepackage{subfiles}
\usepackage{hyperref}       
\usepackage{mathrsfs}
\usepackage{geometry}
\title{\LARGE \bf
Self-supervised Event-based Monocular Depth Estimation using Cross-modal Consistency 
}

\author{Junyu Zhu$^{1}$, Lina Liu$^{1}$, Bofeng Jiang$^{1}$, Feng Wen$^{2}$, Hongbo Zhang$^{2}$, Wanlong Li$^{2*}$ and Yong Liu$^{1*}$
\thanks{$^{1}$Junyu Zhu, Lina Liu, Bofeng Jiang and Yong Liu are with the Institute of Cyber-Systems and Control, Zhejiang University, 
Hangzhou, China. email:\{junyuzhu, linaliu, 22160068\}@zju.edu.cn, yongliu@iipc.zju.edu.cn. }
\thanks{$^{2}$Wanlong Li, Feng Wen and Hongbo Zhang are with Noah’s Ark Lab, Huawei Technologies, Beijing, China. email:\{liwanlong, wenfeng3, zhanghongbo888\}@huawei.com.}
\thanks{$^{*}$Corresponding authors: Wanlong Li and Yong Liu.}
}

\begin{document}

\maketitle
\thispagestyle{empty}
\pagestyle{empty}

\begin{abstract}
An event camera is a novel vision sensor that can capture per-pixel brightness changes and output a stream of asynchronous ``events''. It has advantages over conventional cameras in those scenes with high-speed motions and challenging lighting conditions because of the high temporal resolution, high dynamic range, low bandwidth, low power consumption, and no motion blur. Therefore, several supervised monocular depth estimation from events is proposed to address scenes difficult for conventional cameras. However, depth annotation is costly and time-consuming. In this paper, to lower the annotation cost, we propose a self-supervised event-based monocular depth estimation framework named EMoDepth. EMoDepth constrains the training process using the cross-modal consistency from intensity frames that are aligned with events in the pixel coordinate. Moreover, in inference, only events are used for monocular depth prediction. Additionally, we design a multi-scale skip-connection architecture to effectively fuse features for depth estimation while maintaining high inference speed. Experiments on MVSEC and DSEC datasets demonstrate that our contributions are effective and that the accuracy can outperform existing supervised event-based and unsupervised frame-based methods.


\end{abstract}

\section{Introduction}

Unlike conventional cameras, instead of capturing intensity frames at a fixed rate, event cameras only report brightness changes at the pixel level once they occur. The output of an event camera is a stream of asynchronous events in the format of $(u_{i},t_{i},p_{i})$ that encode the pixel location $u_{i}=(x_{i},y_{i})$, the time $t_{i}$, and the polarity $p_{i}$ that denotes the sign of the brightness change that exceeds a threshold of $\pm C$. Such sensors have several advantages, e.g., very high dynamic range(140 dB vs. 60dB of conventional cameras), high temporal resolution and low latency(both in the order of microseconds), no motion blur, and low power consumption. These advantages give event cameras great potential for machine vision applications in challenging scenes with high-speed motions and high dynamic range. These years, event cameras are attracting the attention of computer vision researchers in various fields, including video reconstruction~\cite{rebecq2019high}, visual odometry~\cite{hidalgo2022event}, optical flow estimation~\cite{gehrig2021raft} and depth estimation~\cite{E2Depth,RAMNet}.

As a task of predicting a dense depth map from a single image, monocular depth estimation is an important and challenging field in computer vision. It helps computers understand the 3D structure of a scene. Thus, it can be applied in various fields such as autonomous driving, augmented reality, and 3D modeling. In the past few years, many monocular depth estimation methods that are based on conventional cameras have been proposed, including supervised methods~\cite{AdaBins,NeWCRFs}, self-supervised methods~\cite{MonoDepth2} and semi-supervised methods~\cite{kuznietsov2017semi,ji2019semi}. Monocular depth estimation using event cameras is relatively less concerned~\cite{Gallego_2018_CVPR,Zhu_2019_CVPR}. However, it is worth excavating the potential capacity of event cameras on depth estimation tasks due to their advantages in those application scenes which need high-frequency depth maps and adaptation to illumination changes. 

\begin{figure}[t]
    \centering
    \begin{minipage}[t]{0.6\linewidth}
        \centerline{\includegraphics[width=\textwidth]{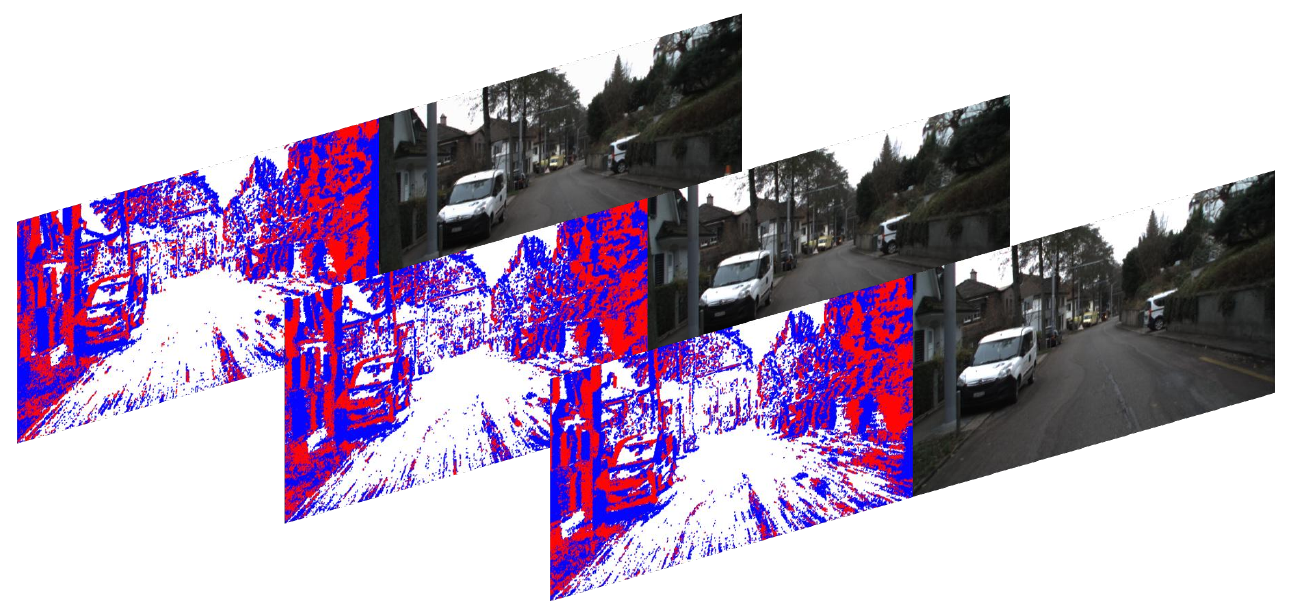}}
        \centerline{(a) Training: Aligned events and intensity frames.}
    \end{minipage}
    \begin{minipage}[t]{1.0\linewidth}
        \centerline{\includegraphics[width=\textwidth]{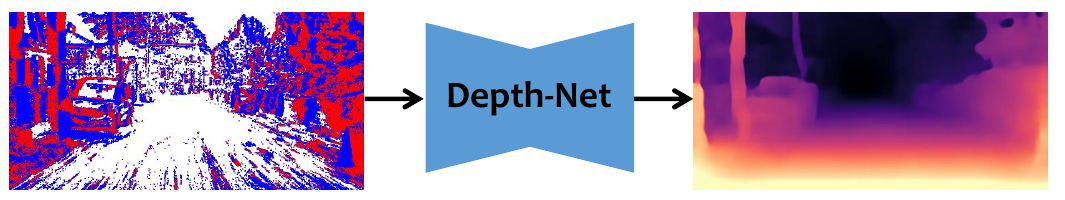}}
        \centerline{(b) Testing: Only events.}
    \end{minipage}
    \caption{\textbf{Method overview.} During the training(a), Pose-Net and Depth-Net are trained with aligned events and intensity frames. During the testing(b), Depth-Net estimates monocular depth map only from events.}\label{overrevie_fig}
    \vspace{-0.4cm}
\end{figure}

Existing supervised event-based monocular depth estimation methods are fed with events~\cite{E2Depth} or a combination of events and frames~\cite{RAMNet} and trained with supervisory from ground truth depth maps collected by extra distance sensors. Expensive annotation cost is the obvious shortcoming of supervised methods. And existing unsupervised event-based monocular depth estimation methods usually depend on self-supervisory from event images deblurring~\cite{Gallego_2018_CVPR,Zhu_2019_CVPR} or photo-consistency between the adjacent event images~\cite{ye2020unsupervised}. However, event data are sparse, so supervisory from events are not dense enough to constraint networks. Furthermore, matching pixels on adjacent event images is difficult because events on corresponding pixels can be very different. 

We propose a framework that consists of Depth-Net and Pose-Net. Depth-Net and Pose-Net are jointly trained with the supervisory signal from the cross-modal consistency of intensity frames that are aligned with events in the pixel coordinate. As shown in Fig.~\ref{overrevie_fig}, the intensity frames are only used for training, and at the testing time, our Depth-Net infers monocular depth maps using events as inputs only. Besides, the decoder of conventional U-Net-based Depth-Net usually fuses upsampling feature maps and original feature maps from the encoder to get multi-scale feature maps for estimating multi-scale depth maps. We design a multi-scale skip-connection architecture based on the finding that Depth-Net can perform better when the above fusion process includes those lower-level feature maps. 

To summarize, our contributions are the following:
\begin{itemize}
  \item We find the photo-consistency of chronological events is weak for self-supervised learning.
  \item We propose a self-supervised framework that exploits cross-modal consistency from intensity frames for event-based monocular depth estimation.
  \item We introduce a multi-scale skip-connection architecture to effectively fuse features for depth estimation.
  \item We demonstrate that our framework can outperform existing supervised event-based methods and unsupervised frame-based methods.
\end{itemize}

\section{Related works}

\begin{figure*}[t]
    \centering
    \includegraphics[scale=0.5]{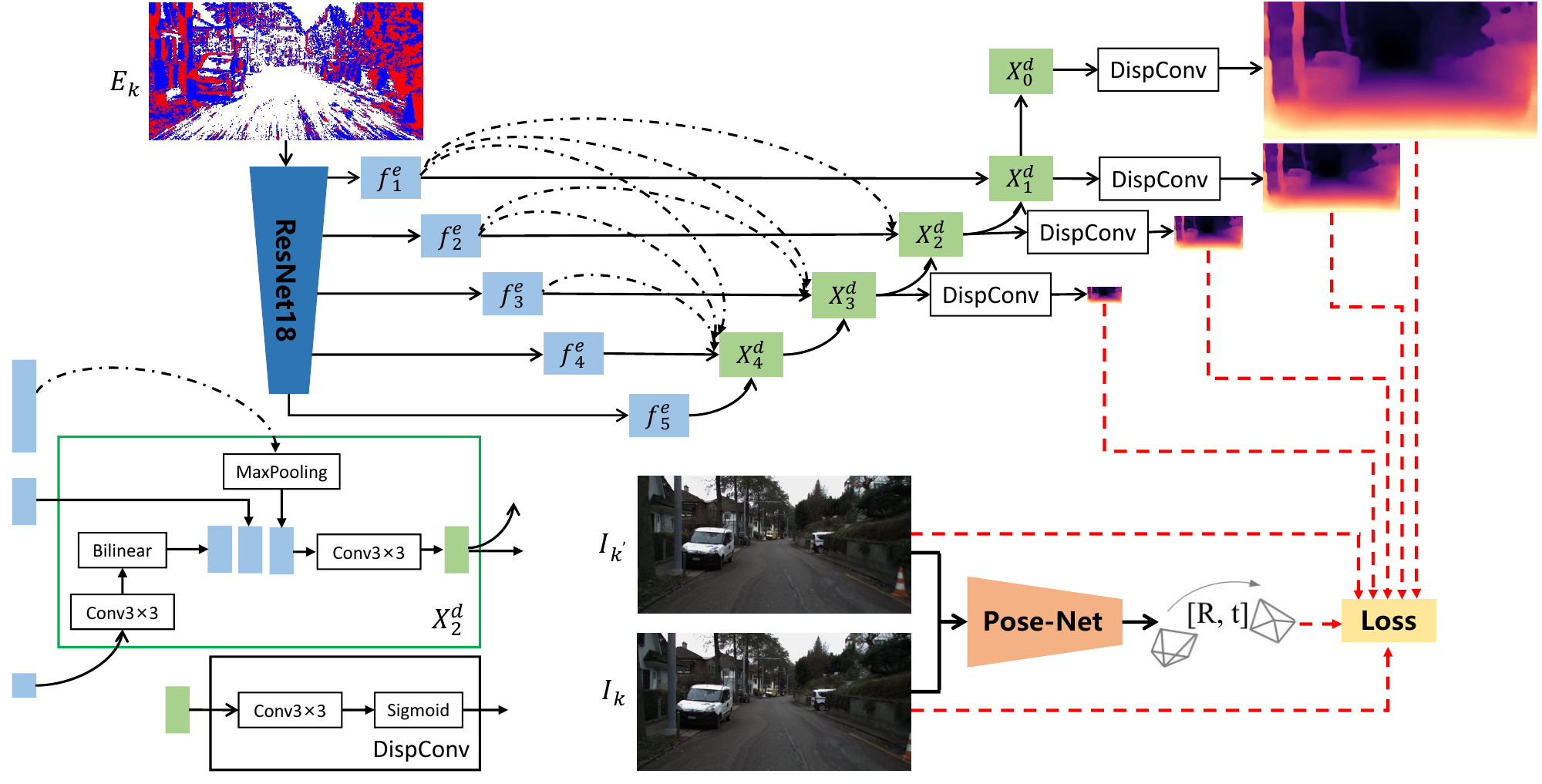}
    \caption{\textbf{Framework illustration.} Our Depth-Net uses ResNet-18 as the encoder and uses several decoder nodes with multi-scale skip-connection architecture as the decoder. Multi-scale features $f^{e}_{i}$ are encoded by ResNet-18 from event voxel grid $E_{k}$. These features are fused by decoder nodes $x^{d}_{i}$ that includes our proposed multi-scale skip-connection architecture, and then outputs of decoder nodes are converted to disparity maps by $DispConv$ blocks. At the same time, the responding intensity frame and adjacent intensity frame are fed to a Pose-Net to predict relative pose $[R|t]$. Finally, the cross-modal consistency loss is computed using multi-scale event-based depth maps, relative pose, and intensity frames. After training, Depth-Net can estimate high-frequency monocular depth only from events.}\label{pipline}
    \vspace{-0.4cm}
\end{figure*}

\subsection{Supervised Monocular Depth Estimation}
Directly using the ground truth depth maps to supervise the training of networks is the most intuitive way to learn the monocular depth. \cite{Eigen} is the first to propose a supervised learning-based method to learn dense monocular depth. They consider monocular depth estimation as a regression task, and later works have followed such an idea for several years. However, \cite{DORN} found that when depth estimation is regarded as a classification task, the depth estimation network can achieve better performance. And in these years, with the emergence of vision transformer~\cite{ViT}, the performance of various visual tasks~\cite{mei2021transvos, detr}, including monocular depth estimation~\cite{AdaBins,NeWCRFs}, has been significantly improved. Supervised monocular depth estimation models usually need a huge amount of parameters to achieve high accuracy, e.g., AdaBins~\cite{AdaBins} has $78M$ parameters, and NeWCRFs~\cite{NeWCRFs} has $270M$ parameters. Furthermore, obtaining GT depth maps at a high cost also limits the application of supervised methods.

\subsection{Self-supervised Monocular Depth Estimation}
To release the burden of collecting ground truth depth maps using expensive distance sensors, e.g., LiDAR, self-supervised monocular depth estimation is proposed to learn monocular depth by exploiting photometric consistency between stereo pairs or monocular sequences. \cite{Garg} is one of the earliest works using stereo pairs to learn monocular depth in a self-supervised manner, and \cite{Monodepth} introduced a left-right consistency loss to produce results comparable to early supervised methods. \cite{SfMLearner} extended such self-supervised framework to monocular sequences using an extra network to predict relative poses between adjacent frames. As a milestone in the self-supervised monocular depth estimation field, \cite{MonoDepth2} helped the accuracy reach new heights by introducing an auto-masking technique and minimum reprojection loss to handle moving objects and occluded areas. To further enhance the accuracy, later works mainly focus on designing more complex network architectures~\cite{xue2020toward,HR_Depth}, using extra semantic constraints~\cite{Edge_Depth,klingner2020self}. Recently, several works~\cite{EPCDepth,RA_Depth} further enhanced the self-supervised method by adopting novel data augmentation approaches to force networks to focus on the key information of images. Also, some works paid attention to the scale ambiguity of self-supervised methods, which are based on monocular sequences and solved the such problem by exploiting prior camera height~\cite{Petrovai_2022_CVPR}, prior object size~\cite{casser2019struct2depth} and linear velocity measurement~\cite{PackNet_SfM}.

\subsection{Event-based Depth Estimation}
Depth estimation based on conventional cameras is prone to be affected by challenging illumination and limited by a fixed frame rate. Therefore, in those scenes with challenging lighting conditions and high-speed motions, depth estimation with event cameras is promising to get more reliable and higher-frequency depth maps. 

Some methods estimate monocular depth using non-learning or unsupervised approaches. \cite{Zhou_2018_ECCV} presented a non-learning semi-dense depth estimation approach for a stereo event camera moving in a static scene. Their method first optimizes an energy function based on the spatiotemporal consistency of events triggered across both stereo image planes simultaneously, then uses a probabilistic fusion strategy to improve density and certainty of estimation. \cite{Gallego_2018_CVPR} proposed a method to estimate monocular depth from events by maximizing the variance of an image of warped events under the limiting assumptions that the camera pose is known and the scene is static. Inspired by \cite{Gallego_2018_CVPR}, \cite{Zhu_2019_CVPR} trained networks in a self-supervised manner to predict optical flow, ego-motion, and depths with the goal of deblurring the event images. 

Also, there are supervised methods for different settings with event cameras. For leveraging the temporal information, \cite{E2Depth} uses a recurrent network for depth estimation from events. \cite{RAMNet} extended this recurrent network for combining asynchronous events and synchronous frames to overcome their demerits and utilize their merits. \cite{cui2022dense} proposed a depth completion network that generates depth maps from sparse events and lidar point clouds.

\section{Method}

\subsection{Method Overview}
In this paper, we aim to train a network that can predict dense monocular depth from a continuous stream of events without ground truth depth maps. To this end, we propose a framework that consists of a Pose-Net and Depth-Net. The input of Depth-Net is event spatiotemporal voxel, which we describe in Sec.~\ref{event_representation}, and the input of Pose-Net is adjacent intensity frames aligned with events. Our framework exploits cross-modal consistency loss, as described in~\ref{loss} from adjacent intensity frames. Additionally, we design a multi-scale skip-connection architecture introduced in~\ref{ms_skip_connection} for better performance of Depth-Net. The pipeline of our framework is shown in Fig.~\ref{pipline}. For the training, we split events into subsequent non-overlapping windows of events $\varepsilon_{k}=\{e_{i}\}^{N-1}_{i=0}$ each spanning a fixed interval $\Delta T=t^{k}_{N-1}-t^{k}_{0}$ where $t^{k}_{N-1}$ is aligned with the time of the intensity frame $I_{k}$ and $\varepsilon_{k}$ is then converted to spatiotemporal voxel $E_{k}$. After training, Depth-Net can infer high-frequency dense depth maps only from events within a fixed interval.


\subsection{Event Representation}\label{event_representation}
The output of an event camera is a stream of asynchronous events, and each event $e_{i}=(u_{i},t_{i}.p_{i})$ records the pixel location $u_{i}$, time $t_{i}$ and polarity $p_{i}$ of per-pixel brightness change. To feed a batch of events within the time window $\Delta T$ to networks, we convert the events to a tensor-like format $E_{k}$ with a fixed dimension. There are several methods~\cite{Zhu-RSS-18,benosman2013event,zihao2018unsupervised,sironi2018hats,gehrig2019end} proposed to represent events. In this paper, for a fair comparison with the majority of learning-based monocular depth estimation methods for events, we represent events as a spatiotemporal voxel grid with a dimension of $B\times H\times W$ that expresses events within the time window $\Delta T$ as $B$ temporal bins. The converting process can be formulated as follows:
\begin{equation}
    E_{k}(u_{k},t_{n})=\sum_{u_{i}=u_{k}}p_{i}max(0,1-|t_{n}-t^{*}_{i}|)
\end{equation}
where $u_{k}$ denotes pixel coordinate on the image plane of $H\times W$, $t_{n}$ belonging to $[0, B-1]$ denotes order number of temporal bins, $t^{*}_{i}=\frac{B-1}{\Delta T}(t_{i}-t_{0})$ is the normalized timestamp of event $e_{i}$ and $u_{i}$ is the pixel location of event $e_{i}$. In this paper, we represent events within $\Delta T=50ms$ as a spatiotemporal voxel grid with $B=5$. 

\subsection{Cross-modal Consistency Loss}\label{photometric_loss}
\begin{figure}[h]
    \centering
    \raisebox{-0.1cm}{\includegraphics[width=8cm]{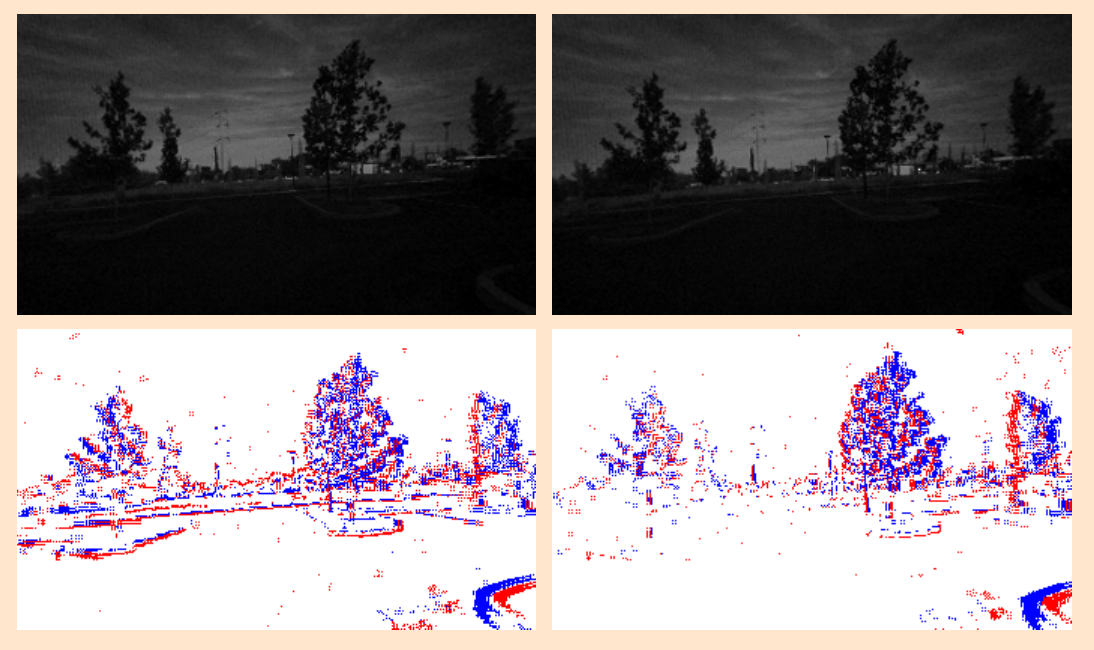}}
    \subfigure{
        \begin{minipage}[t]{0.22\textwidth}
            \centering
            {\scriptsize{(a)}}
        \end{minipage}
        \begin{minipage}[t]{0.22\textwidth}
            \centering
            {\scriptsize{(b)}}
        \end{minipage}
    }\vspace{-2mm}
    \caption{\textbf{Adjacent event spatiotemporal voxels visualization.} The first row shows grayscale intensity frames from MVSEC, and the second row shows aligned events. We sum the voxel along the channel axis, then use blue and red pixels to represent negative and positive results. (a) and (b) are respectively from sample 125 and sample 126 of sequence $outdoor\_day1$. As shown in the figure, corresponding pixels can have very different events. So, the photoconsistency of event spatiotemporal voxel is weak.}\label{events_comparison}
\end{figure}

For constraining Pose-Net and Depth-Net in a self-supervised manner, the very intuitive idea is to exploit the photoconsistency between adjacent event spatiotemporal voxels. However, we found the photoconsistency is unsuitable for events in the adjacent time windows (see Fig.~\ref{events_comparison}) because brightness changes on corresponding pixels can have contrary polarity and very different normalized timestamps that are closely related to camera motion. Moreover, the event spatiotemporal voxel is not dense enough (<15\% in the MVSEC dataset) to provide networks with intensive self-supervisory signals.

As a solution, we choose to get consistency signals from other modals, e.g. intensity frames that are aligned with events in the pixel coordinate. With the depth estimation $D_{k}$ from an event spatiotemporal voxel $E_{k}$ and the pose estimation $T_{k\rightarrow k^{'}}$ between the corresponding intensity frame $I_{k}$ and the adjacent intensity frame $I_{k^{'}}$, a synthesized intensity frame $I^{k^{'}}_{k}$ can be reconstructed from frame $I_{k^{'}}$ with depth $D_{k}$ and pose $T_{k\rightarrow k^{'}}$ using inverse-warping~\cite{SfMLearner}. Then, combining SSIM~\cite{SSIM} loss and L1 loss, re-projection error $pe(I_{k},I^{k^{'}}_{k})$ can be computed as following:
\begin{equation}
    \begin{aligned}
    pe(I_{k},I^{k^{'}}_{k})=&\frac{\alpha}{2}(1-SSIM(I_{k},I^{k^{'}}_{k}))\\
    &+(1-\alpha)\|I_{k}-I^{k^{'}}_{k}\|_{1}
    \end{aligned}
\end{equation}
where SSIM is used to measure the structural similarity of images and computed over a $3\times 3$ pixel window, and $\alpha$ is set to be 0.85. We follow the per-pixel minimum reprojection loss introduced by~\cite{MonoDepth2} to handle occlusion. The cross-modal consistency loss $L_{p}$ can be formulated as:
\begin{equation}
    L_{cc}=\min_{k^{'}}pe(I_{k},I^{k^{'}}_{k})
\end{equation}
where $k^{'}\in \{k-1,k+1\}$ thus two frames temporally adjacent to $I_{k}$ are used as source frames.

\subsection{Auto-Masking}
To mask those pixels that remain the same due to a relatively static state and low texture, we apply the auto-masking technique proposed in \cite{MonoDepth2}. The non-static mask $M_{ns}$ is defined as:
\begin{equation}
    M_{ns}=[\min_{k^{'}}pe(I_{k},I^{k^{'}}_{k})<\min_{k^{'}}pe(I_{k},I_{k^{'}})]
\end{equation}
where [] is the Iverson bracket.

\subsection{Training Loss}\label{loss}
We combine cross-modal consistency loss $L_{cc}$ and non-static mask $M_{ns}$ as $L=M_{ns}L_{cc}$, and average over pixels and scales to get final training loss $L$:
\begin{equation}
    L=\frac{1}{s}\sum_{j}^{s}(\frac{1}{T}\sum_{i}^{T}M^{j}_{ns}(p_{i})L^{j}_{cc}(p_{i}))
\end{equation}
where $s$ denotes the number of scales, $p_{i}$ demotes a pixel coordinate and $T$ denotes the number of pixels of an image.

\subsection{Multi-scale Skip-connection}\label{ms_skip_connection}
U-Net-based architecture has become a common choice~\cite{MonoDepth2,DIFFNet} for designing the decoder of Depth-Net in the self-supervised monocular depth estimation field. As one of the core components of U-Net, skip-connection is used for recovering information lost in the downsampling process~\cite{HR_Depth}. For an event spatiotemporal voxel, due to its natural sparsity (<15\% in the MVSEC), recovering information lost becomes essential. Moreover, we think skip-connection from the encoder feature with the same level is insufficient. Inspired by \cite{huang2020unet}, we propose a multi-scale skip-connection that connects with the feature of the same level and the lower-level features.

Let $f^{e}_{i}$ denote a feature from the encoder, $x^{d}_{i}$ denote the output of decoder node $X^{d}_{i}$. The $x^{d}_{i}$ can be computed as:
\begin{equation}
    x^{d}_{i}=
    \begin{cases}
    D(U(x^{d}_{i+1})),& i=0 \\
    D([[f^{e}_{i},[M(f^{e}_{k})]^{i-1}_{k=1}],U(x^{d}_{i+1})]),& i>0 
    \end{cases}
\end{equation}
where $U(\cdot)$ denotes an upsampling block consists of $3\times3$ convolution layer with ELU activation and bilinear interpolation. $M(\cdot)$ is a maxpooling layer to downsample lower-level features, $D(\cdot)$ is a feature fusion block made up of $3\times3$ convolution layer with ELU activation and $[\cdot]$ denotes the concatenation operation. The details of multi-scale skip-connection can be seen in Fig.~\ref{pipline}.

With such designing, lower-level features can directly participate in the feature fusion process. Thus, information is less lost, and Depth-Net can achieve better performance.

\section{Experiments}

In this section, we evaluate our proposed framework on the MVSEC dataset to present its qualitative and quantitative results and compare them with previous works. Also, ablation studies on the MVSEC dataset are conducted to demonstrate the effectiveness of our contributions. Further, to test our framework on higher resolution events and color intensity frames in more scenes, we evaluate the framework on the DSEC dataset, a very new event camera dataset.

\subsection{Datasets}
\textbf{MVSEC.} MVSEC\footnote{\url{https://daniilidis-group.github.io/mvsec/}} is the most commonly used dataset for the event-based depth estimation task. It was captured by a synchronized stereo pair event-based camera system carried on a handheld rig, flown by a hexacopter, driven on top of a car, and mounted on a motorcycle in different scenes and at various illumination levels. 

\textbf{DSEC.} DSEC is a recently proposed dataset that offers data from a wide-baseline stereo setup of two color frame cameras with a resolution of $1080\times 1440$ and two monochrome event cameras with a resolution of $480\times 640$, and a lidar. DSEC is in driving scenarios and contains 53 sequences in different illumination conditions.

\begin{table*}[t]
\centering
\caption{\textbf{Quantitative results on the MVSEC dataset.} Comparison of the average absolute depth error(in meters) at different cutoff distance. $\mathbb{I}$ means using instensity frames as input. $\mathbb{E}$ means using events as input. And $\mathbb{I+E}$ means using both intensity frames and events as input.} 
\label{mvsec_quantitative_results}
\scriptsize
\setlength
\tabcolsep{3pt}
{
\setlength{\arrayrulewidth}{.1em}
\begin{tabular}{c|c|c|c|c|c|cc}
\hline
\multicolumn{1}{c|}{\multirow{2}{*}{\textbf{Dataset}}} & \multicolumn{1}{c|}{\multirow{2}{*}{\textbf{Cutoff}}} & \multicolumn{1}{c|}{\textbf{Supervised hybrid}} & \multicolumn{1}{c|}{\textbf{Supervised frame-based}} & \multicolumn{1}{c|}{\textbf{Unsupervised frame-based}} & \multicolumn{1}{c|}{\textbf{Supervised event-based}} & \multicolumn{2}{c}{\textbf{Unsupervised event-based}} \\
\cline{3-8}
\multicolumn{1}{c|}{} & \multicolumn{1}{c|}{} & \multicolumn{1}{c|}{RAM-Net($\mathbb{I+E}$)~\cite{RAMNet}} & \multicolumn{1}{c|}{RAM-Net($\mathbb{I}$)~\cite{RAMNet}} & \multicolumn{1}{c|}{Ours($\mathbb{I}$)} & \multicolumn{1}{c|}{E2Depth~\cite{E2Depth}} & \multicolumn{1}{c}{Zhu et al.~\cite{Zhu_2019_CVPR}} & \multicolumn{1}{c}{Ours($\mathbb{E}$)} \\
\hline
               & 10m & \textbf{1.39} & 1.74 & 1.54 & 1.85 & 2.72 & 1.40 \\
outdoor day1   & 20m & 2.17 & 2.55 & 2.23 & 2.64 & 3.84 & \textbf{2.07} \\
               & 30m & 2.76 & 3.07 & 2.71 & 3.13 & 4.40 & \textbf{2.65} \\ 
\hline
               & 10m & 2.50 & 2.72 & 3.24 & 3.38 & 3.13 & \textbf{2.18} \\
outdoor night1 & 20m & 3.19 & 3.35 & 3.74 & 3.82 & 4.02 & \textbf{2.70} \\
               & 30m & 3.82 & 3.99 & 4.60 & 4.46 & 4.89 & \textbf{3.64} \\ 
\hline
               & 10m & \textbf{1.21} & 1.36 & 3.16 & 1.67 & 2.19 & 2.06 \\
outdoor night2 & 20m & \textbf{2.31} & 2.42 & 3.65 & 2.63 & 3.15 & 2.76 \\
               & 30m & \textbf{3.28} & 3.47 & 4.24 & 3.58 & 3.92 & 3.42 \\ 
\hline
               & 10m & \textbf{1.01} & 1.20 & 3.09 & 1.42 & 2.86 & 2.09 \\
outdoor night3 & 20m & 2.34 & 2.44 & 3.55 & \textbf{2.33} & 4.46 & 2.82 \\
               & 30m & 3.43 & 3.64 & 4.20 & \textbf{3.18} & 5.05 & 3.52 \\ 
\hline
\end{tabular}
}\vspace{-1ex}
\end{table*}

\begin{table*}[h]
\centering
\caption{\textbf{Quantitative results on the DSEC dataset.} $\mathbb{I}$ means using instensity frames as input and $\mathbb{E}$ means using events as input. Results in last three columns denote average absolute depth errors (in meters) at different maximum cut-off depths.}
\label{dsec_quantitative_results}
\scriptsize
\setlength\tabcolsep{3pt}
\begin{tabular}{c|ccccc|ccc|ccc}
    \toprule
        \textbf{Method} & \textbf{Abs Rel}$\downarrow$ &  \textbf{Sq Rel} $\downarrow$ &\textbf{RMSE}$\downarrow$ &  \textbf{RMSE log}$\downarrow$ & \textbf{SI log}$\downarrow$ & $ \boldsymbol{\delta < 1.25\uparrow}$ & $\boldsymbol{\delta < 1.25^2 \uparrow}$ & $\boldsymbol{\delta < 1.25^3 \uparrow}$ & $\boldsymbol{c=10 \downarrow}$ & $\boldsymbol{c=20 \downarrow}$  & $\boldsymbol{c=30 \downarrow}$\\ 
    \toprule
    Ours($\mathbb{I}$) & 0.166 & 1.346 & 5.934 & 0.229 & 0.050 & 0.749 & 0.934 & 0.980 & 1.042 & 1.942 & 2.622 \\
    Ours($\mathbb{E}$) & \textbf{0.142} & \textbf{1.152} & \textbf{5.258} & \textbf{0.202} & \textbf{0.043} & \textbf{0.819} & \textbf{0.948} & \textbf{0.983} & \textbf{0.960} & \textbf{1.550} & \textbf{2.205} \\
    \bottomrule
\end{tabular}
\end{table*}

\begin{figure*}[t]
\centering
    \subfigure{
        \begin{minipage}[t]{0.14\linewidth}
            \centering
            \raisebox{-0.15cm}{\includegraphics[width=2.5cm]{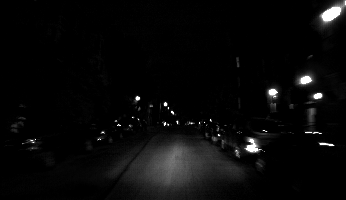}}
        \end{minipage}
        \begin{minipage}[t]{0.14\linewidth}
            \centering
            \raisebox{-0.15cm}{\includegraphics[width=2.5cm]{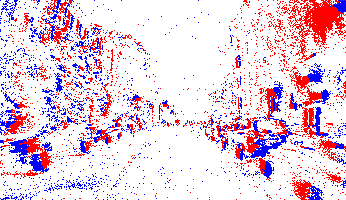}}
        \end{minipage}
        \begin{minipage}[t]{0.14\linewidth}
            \centering
            \raisebox{-0.15cm}{\includegraphics[width=2.5cm]{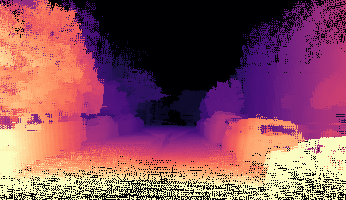}}
        \end{minipage}
        \begin{minipage}[t]{0.14\linewidth}
            \centering
            \raisebox{-0.15cm}{\includegraphics[width=2.5cm]{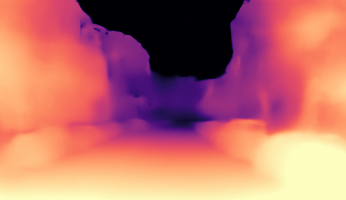}}
        \end{minipage}
        \begin{minipage}[t]{0.14\linewidth}
            \centering
            \raisebox{-0.15cm}{\includegraphics[width=2.5cm]{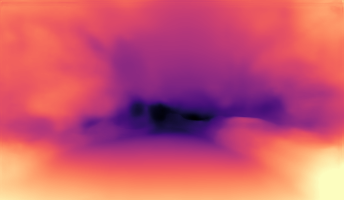}}
        \end{minipage}
        \begin{minipage}[t]{0.14\linewidth}
            \centering
            \raisebox{-0.15cm}{\includegraphics[width=2.5cm]{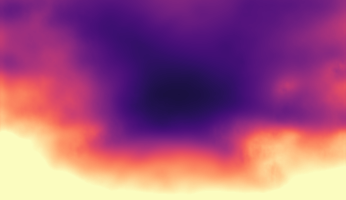}}
        \end{minipage}
    }\vspace{-2mm}
    \subfigure{
        \begin{minipage}[t]{0.14\linewidth}
            \centering
            \raisebox{-0.15cm}{\includegraphics[width=2.5cm]{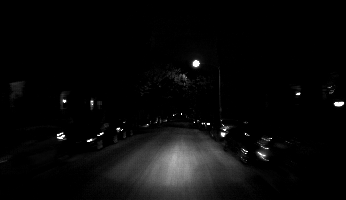}}
        \end{minipage}
        \begin{minipage}[t]{0.14\linewidth}
            \centering
            \raisebox{-0.15cm}{\includegraphics[width=2.5cm]{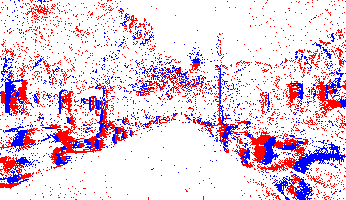}}
        \end{minipage}
        \begin{minipage}[t]{0.14\linewidth}
            \centering
            \raisebox{-0.15cm}{\includegraphics[width=2.5cm]{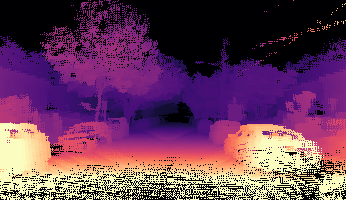}}
        \end{minipage}
        \begin{minipage}[t]{0.14\linewidth}
            \centering
            \raisebox{-0.15cm}{\includegraphics[width=2.5cm]{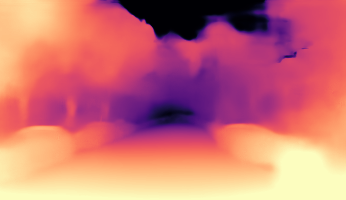}}
        \end{minipage}
        \begin{minipage}[t]{0.14\linewidth}
            \centering
            \raisebox{-0.15cm}{\includegraphics[width=2.5cm]{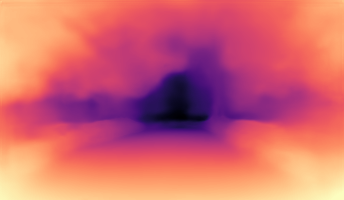}}
        \end{minipage}
        \begin{minipage}[t]{0.14\linewidth}
            \centering
            \raisebox{-0.15cm}{\includegraphics[width=2.5cm]{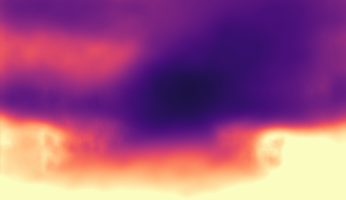}}
        \end{minipage}
    }\vspace{-2mm}
    \subfigure{
        \begin{minipage}[t]{0.14\textwidth}
            \centering
            {\scriptsize{Frame}}
        \end{minipage}
        \begin{minipage}[t]{0.14\textwidth}
            \centering
            {\scriptsize{Events}}
        \end{minipage}
        \begin{minipage}[t]{0.14\textwidth}
            \centering
            {\scriptsize{GT Depth}}
        \end{minipage}
        \begin{minipage}[t]{0.14\textwidth}
            \centering
            {\scriptsize{E2Depth~\cite{E2Depth}}}
        \end{minipage}
        \begin{minipage}[t]{0.14\textwidth}
            \centering
            {\scriptsize{RAMNet($\mathbb{I+E}$)~\cite{RAMNet}}}
        \end{minipage}
        \begin{minipage}[t]{0.14\textwidth}
            \centering
            {\scriptsize{Ours($\mathbb{E}$)}}
        \end{minipage}
    }\vspace{-3mm}
    \caption{\textbf{Qualitative results on the MVSEC dataset.} The qualitative result of Zhu et al.~\cite{Zhu_2019_CVPR} is omitted because their code isn't publicly available. Our EMoDepth has a relatively more reasonable prediction on distant areas, e.g., trees and sky in the upper part of the image.}\label{mvsec_qualitative_result}
\end{figure*}

\begin{figure*}[t]
\centering
    \subfigure{
        \begin{minipage}[t]{0.17\linewidth}
            \centering
            \raisebox{-0.15cm}{\includegraphics[width=3cm]{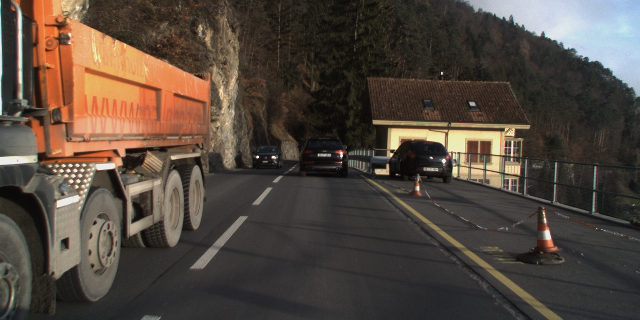}}
        \end{minipage}
        \begin{minipage}[t]{0.17\linewidth}
            \centering
            \raisebox{-0.15cm}{\includegraphics[width=3cm]{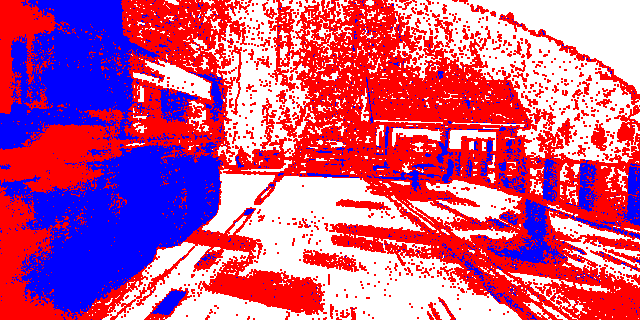}}
        \end{minipage}
        \begin{minipage}[t]{0.17\linewidth}
            \centering
            \raisebox{-0.15cm}{\includegraphics[width=3cm]{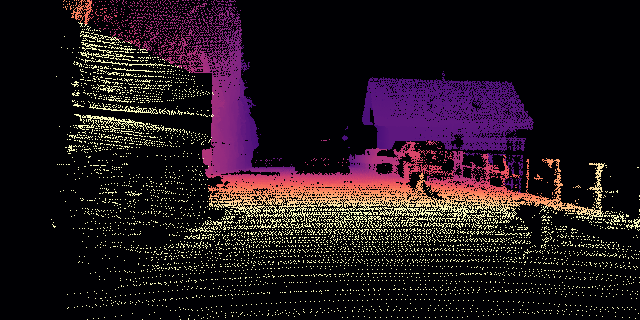}}
        \end{minipage}
        \begin{minipage}[t]{0.17\linewidth}
            \centering
            \raisebox{-0.15cm}{\includegraphics[width=3cm]{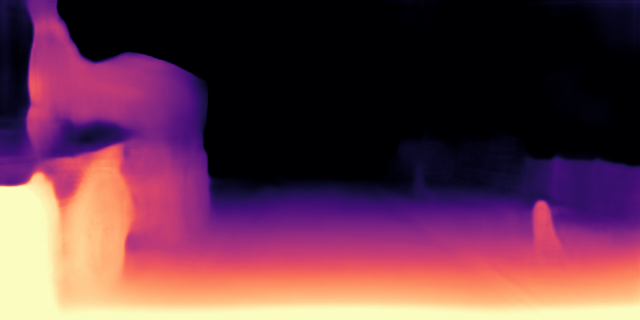}}
        \end{minipage}
        \begin{minipage}[t]{0.17\linewidth}
            \centering
            \raisebox{-0.15cm}{\includegraphics[width=3cm]{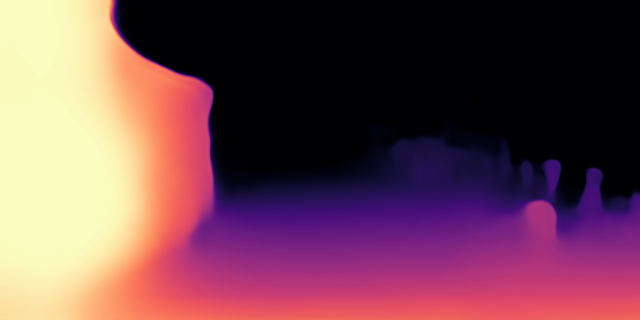}}
        \end{minipage}
    }\vspace{-2mm}
    \subfigure{
        \begin{minipage}[t]{0.17\linewidth}
            \centering
            \raisebox{-0.15cm}{\includegraphics[width=3cm]{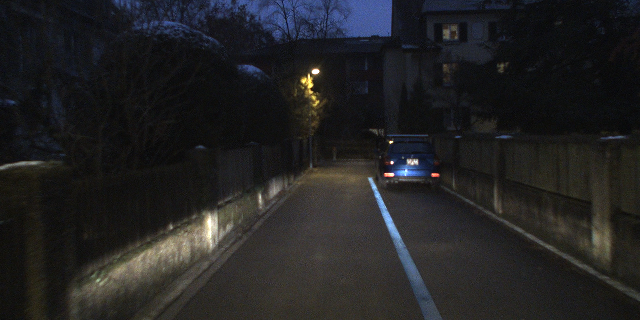}}
        \end{minipage}
        \begin{minipage}[t]{0.17\linewidth}
            \centering
            \raisebox{-0.15cm}{\includegraphics[width=3cm]{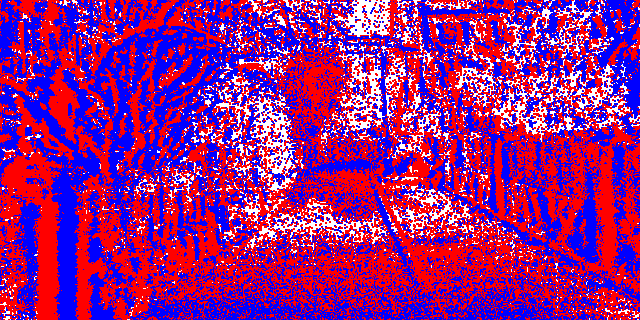}}
        \end{minipage}
        \begin{minipage}[t]{0.17\linewidth}
            \centering
            \raisebox{-0.15cm}{\includegraphics[width=3cm]{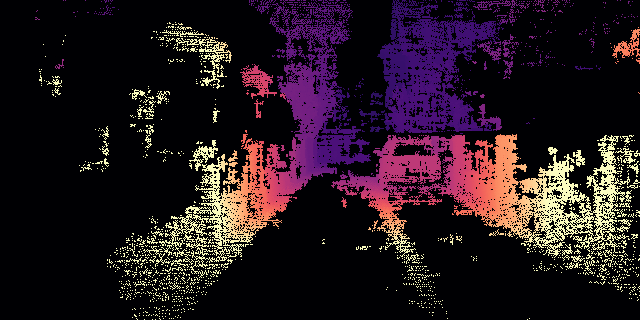}}
        \end{minipage}
        \begin{minipage}[t]{0.17\linewidth}
            \centering
            \raisebox{-0.15cm}{\includegraphics[width=3cm]{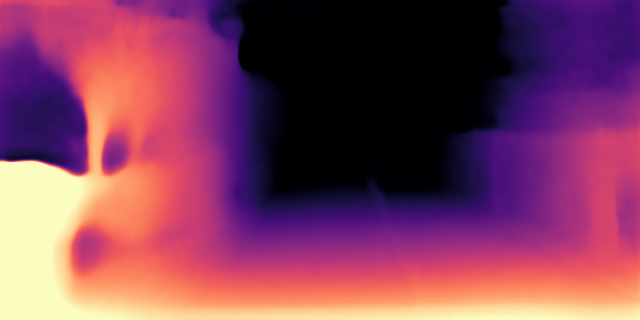}}
        \end{minipage}
        \begin{minipage}[t]{0.17\linewidth}
            \centering
            \raisebox{-0.15cm}{\includegraphics[width=3cm]{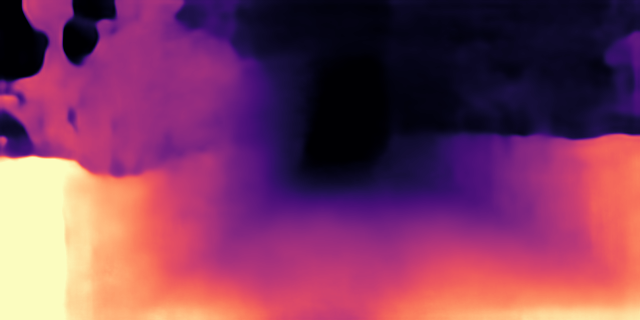}}
        \end{minipage}
    }\vspace{-2mm}
    \subfigure{
        \begin{minipage}[t]{0.17\linewidth}
            \centering
            \raisebox{-0.15cm}{\includegraphics[width=3cm]{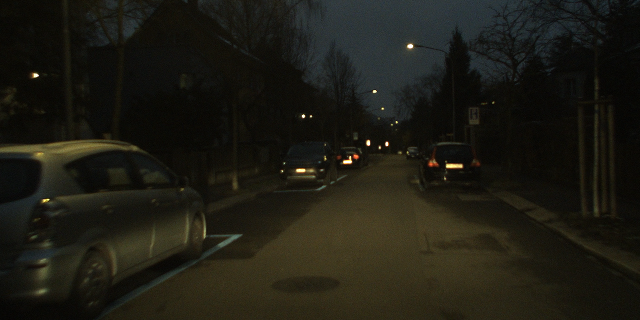}}
        \end{minipage}
        \begin{minipage}[t]{0.17\linewidth}
            \centering
            \raisebox{-0.15cm}{\includegraphics[width=3cm]{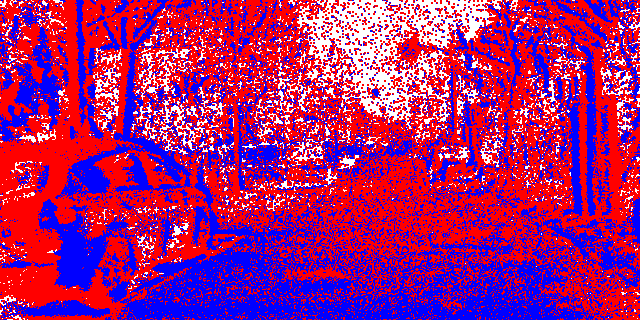}}
        \end{minipage}
        \begin{minipage}[t]{0.17\linewidth}
            \centering
            \raisebox{-0.15cm}{\includegraphics[width=3cm]{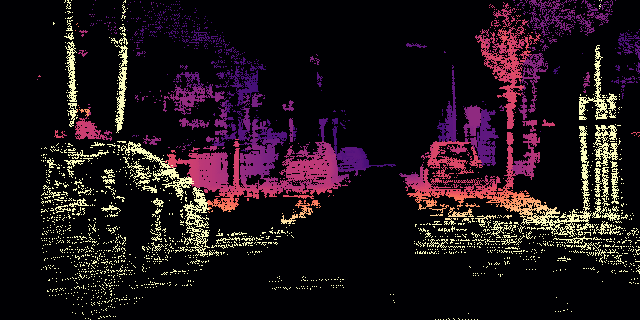}}
        \end{minipage}
        \begin{minipage}[t]{0.17\linewidth}
            \centering
            \raisebox{-0.15cm}{\includegraphics[width=3cm]{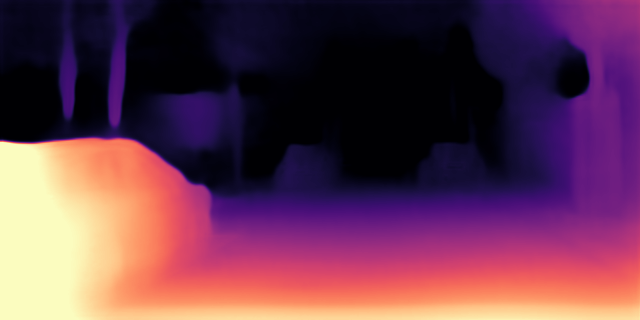}}
        \end{minipage}
        \begin{minipage}[t]{0.17\linewidth}
            \centering
            \raisebox{-0.15cm}{\includegraphics[width=3cm]{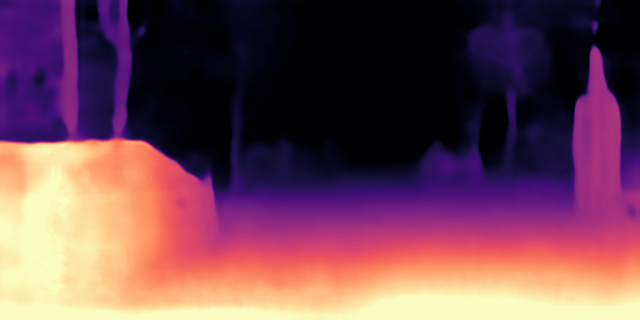}}
        \end{minipage}
    }\vspace{-2mm}
    \subfigure{
        \begin{minipage}[t]{0.17\textwidth}
            \centering
            {\scriptsize{Frame}}
        \end{minipage}
        \begin{minipage}[t]{0.17\textwidth}
            \centering
            {\scriptsize{Events}}
        \end{minipage}
        \begin{minipage}[t]{0.17\textwidth}
            \centering
            {\scriptsize{GT Depth}}
        \end{minipage}
        \begin{minipage}[t]{0.17\textwidth}
            \centering
            {\scriptsize{Ours($\mathbb{I}$)}}
        \end{minipage}
        \begin{minipage}[t]{0.17\textwidth}
            \centering
            {\scriptsize{Ours($\mathbb{E}$)}}
        \end{minipage}
    }\vspace{-3mm}
    \caption{\textbf{Qualitative results on the DSEC dataset.} $\mathbb{I}$ means using instensity frames as input and $\mathbb{E}$ means using events as input.}\label{dsec_qualitative_result}
    \vspace{-0.4cm}
\end{figure*}

\subsection{Inplementation Details}\label{inplementation_details}
\textbf{Network architectures.} We use the same Pose-Net as previous works~\cite{MonoDepth2}. Moreover, we revise the first convolution layers of Depth-Net and Pose-Net to fit the channel number of events voxel and grayscale intensity frame. Considering different distributions of depth, the output $\sigma$ of the Depth-Net is further constrained between 0.1 and 100 units for the MVSEC dataset and 0.1 and 60 units for the DSEC dataset with $D=1/(a\sigma+b)$. For the MVSEC dataset, we take depth estimation of 4 scales for training, while only the maximum scale depth estimation is used for the DSEC dataset.

\textbf{Data preprocessing of MVSEC.} For the MVSEC dataset, we use the split proposed in \cite{E2Depth}. More specifically, we train our networks on $outdoor\_day2$ sequence that is splited into 8523 training samples, 1826 validation samples, and 1826 testing samples. After training, we evaluate networks on other outdoor sequences. Considering static frames can disturb self-supervised training, we follow Zhou $et\; al.$'s~\cite{SfMLearner} pre-processing and get 6817 training samples. To get a suitable input size, we improve resolution from the original $260\times 346$ to larger $288\times 352$ using zero padding. An evaluation is done on the upper middle areas with a resolution of $200\times 346$.

\textbf{Data preprocessing of DSEC.} For the DSEC dataset, there are only optical flow benchmark and stereo matching benchmark on the official website\footnote{\url{https://dsec.ifi.uzh.ch/}} and the ground truth disparity maps of official testing sequences are reserved by the official server. For convenience, we split official training sequences into 28 for training and 13 for testing. The training set containing 16838 images has all images per training scene, and the testing set containing 1300 images has 100 images per testing scene. In the DSEC dataset, events and intensity frames are separately collected by different synchronized sensors. Following~\cite{mostafavi2021event}, we warp the rectified intensity frames to the event locations according to the calibration parameters provided. First, we get undistort and aligned event spatiotemporal voxels with intensity frames at a resolution of $480\times 640$. Then, these data are center-cropped to $320\times 640$ for cropping the car-hood.

\textbf{Data augmentation.} We randomly flip the input images horizontally and apply color augmentations with a probability of 50\%. For the color augmentation, we perform random brightness, contrast, saturation, and hue jitter by sampling uniform distributions in ranges of [0.8,1.2], [0.8,1.2], [0.8,1.2], [0.9,1.1], respectively. Note that color augmentation is only applied to the DSEC dataset.

\textbf{Training setup.} Our work is implemented in PyTorch on a single GTX1080Ti with 11GB memory. We train the networks for 10 epochs with Adam optimizer ($\beta_{1}$ = 0.9, $\beta_{2}$ = 0.999) and a batch size of 8. The initial learning rate is $1\times 10^{-4}$ for the first 8 epochs and $1\times 10^{-5}$ for the remaining.

\subsection{Depth Estimation Results}
We compare our framework performance with other methods on the MVSEC dataset. As in previous works, we report the average mean errors at maximum cutoff depths of 10m, 20m, and 30m. The quantitative and qualitative results are represented in Tab.~\ref{mvsec_quantitative_results} and Fig.~\ref{mvsec_qualitative_result}, respectively. The quantitative results demonstrate that our EMoDepth outperforms unsupervised frame-based methods and even outperforms RAMNet($\mathbb{I}$) and E2Depth, which design recurrent architectures to leverage the temporal information and use additional synthetic data for better results. Due to the more practical constraints of proposed cross-modal consistency, EmoDepth performs better than Zhu et al.~\cite{Zhu_2019_CVPR}, which utilizes relatively weaker supervision by deblurring the event images. The qualitative results show that compared with E2Depth and RAMNet, our EMoDepth has a relatively more reasonable prediction on distant areas, e.g., trees and sky in the upper part of the image.

Also, we test our EMoDepth on the DSEC. Since there are no event-based monocular depth estimation methods reporting results on the DSEC dataset, we test our EMoDepth using frames($\mathbb{I}$) as input and using events($\mathbb{E}$) as input for comparison. The quantitative and qualitative results represented in Tab.~\ref{dsec_quantitative_results} and Fig.~\ref{dsec_qualitative_result} also show that when using the event as input, our EMoDepth can achieve better performance. Results also indicate that performance on the DSEC dataset is significantly better than on the MVSEC dataset. It can be explained as the benefits of denser events and color intensity frames.

\subsection{Ablation studies}

\paragraph{Choice of self-supervisory signal}
We propose to exploit aligned intensity frames to form self-supervisory signals. To validate its effectiveness, we directly use the consistency of adjacent event spatiotemporal voxels to train the networks. As shown in Tab.~\ref{mvsec_sss_ablation}, the accuracy drops dramatically when the self-supervisory signals come from event spatiotemporal voxels. This is consistent with our analysis in~\ref{photometric_loss}.

\begin{table}[h]
\centering
\caption{\textbf{Ablation studied of different self-supervisory signals on MVSEC dataset}} 
\label{mvsec_sss_ablation}
\scriptsize
\setlength{\arrayrulewidth}{.1em}
\begin{tabular}{c|c|cc}
\hline
\multicolumn{1}{c|}{\multirow{2}{*}{\textbf{Dataset}}} & \multicolumn{1}{c|}{\multirow{2}{*}{\textbf{Cutoff}}} & \multicolumn{2}{c}{\textbf{Self-supervisory signal}} \\
\cline{3-4}
\multicolumn{1}{c|}{} & \multicolumn{1}{c|}{} & \multicolumn{1}{c}{\textbf{Event spatiotemporal voxels}} & \multicolumn{1}{c}{\textbf{Intensity frames}} \\
\hline
               & 10m & 3.90 & \textbf{1.40} \\
outdoor day1   & 20m & 3.79 & \textbf{2.07} \\
               & 30m & 4.89 & \textbf{2.65} \\ 
\hline
               & 10m & 5.55 & \textbf{2.18} \\
outdoor night1 & 20m & 4.57 & \textbf{2.70} \\
               & 30m & 5.72 & \textbf{3.64} \\ 
\hline
               & 10m & 5.76 & \textbf{2.06} \\
outdoor night2 & 20m & 4.48 & \textbf{2.76} \\
               & 30m & 5.26 & \textbf{3.42} \\ 
\hline
               & 10m & 5.87 & \textbf{2.09} \\
outdoor night3 & 20m & 4.39 & \textbf{2.82} \\
               & 30m & 5.33 & \textbf{3.52} \\ 
\hline
\end{tabular}
\end{table}

\paragraph{Effect of multi-scale skip-connection}
We introduce multi-scale skip-connection to reduce information loss of sparse events. And the results in \ref{mvsec_sc_ablation} show that Depth-Net can achieve better performance when the multi-scale skip-connection is introduced when compared to the baseline (MonoDepth2). 

\begin{table}[h]
    \centering
    \caption{\textbf{Ablation studied of multi-scale skip-connection on MVSEC dataset.}} 
    \label{mvsec_sc_ablation}
    \scriptsize
    \setlength{\arrayrulewidth}{.1em}
    \begin{tabular}{c|c|cc}
    \hline
    \multicolumn{1}{c|}{\textbf{Dataset}} & \multicolumn{1}{c|}{\textbf{Cutoff}} & \multicolumn{1}{c}{\textbf{Baseline}} & \multicolumn{1}{c}{\textbf{+ multi-scale SC}} \\
    \hline
                   & 10m & 1.48 & \textbf{1.40~($\downarrow$0.08)} \\
    outdoor day1   & 20m & 2.22 & \textbf{2.07~($\downarrow$0.15)} \\
                   & 30m & 2.74 & \textbf{2.65~($\downarrow$0.09)} \\ 
    \hline
                   & 10m & 2.55 & \textbf{2.18~($\downarrow$0.37)} \\
    outdoor night1 & 20m & 3.06 & \textbf{2.70~($\downarrow$0.36)} \\
                   & 30m & 3.95 & \textbf{3.64~($\downarrow$0.31)} \\ 
    \hline
                   & 10m & 2.47 & \textbf{2.06~($\downarrow$0.41)} \\
    outdoor night2 & 20m & 3.03 & \textbf{2.76~($\downarrow$0.27)} \\
                   & 30m & 3.64 & \textbf{3.42~($\downarrow$0.22)} \\ 
    \hline
                   & 10m & 2.44 & \textbf{2.09~($\downarrow$0.35)} \\
    outdoor night3 & 20m & 2.95 & \textbf{2.82~($\downarrow$0.13)} \\
                   & 30m & 3.60 & \textbf{3.52~($\downarrow$0.08)} \\ 
    \hline
    \end{tabular}
\end{table}

\paragraph{Time consumption}
We also test the inference time of E2Depth, RAM-Net, and our EMoDepth, on the MVSEC with a GTX1080TI GPU. The results in  Tab.~\ref{time_consummption} show that the inference speed of EMoDepth is significantly faster than E2Depth and RAM-Net. And the introduction of multi-scale skip-connnection does not affect real-time performance much (from 4.67ms to 5.25ms) while improving performance. 

\begin{table}[h]
    \centering
    \caption{\textbf{Time consumption on MVSEC dataset using different networks.}} 
    \label{time_consummption}
    \scriptsize
    \setlength{\arrayrulewidth}{.1em}
    \begin{tabular}{c|c}
    \hline
    \multicolumn{1}{c|}{\textbf{Network}} & \multicolumn{1}{c}{\textbf{Time consomption (ms)}} \\
    \hline
    E2Depth & 24.22 \\
    RAM-Net & 25.68 \\
    EMoDepth(w/o multi-scale SC) & \textbf{4.67} \\ 
    EMoDepth & 5.25 \\
    \hline
    \end{tabular}
\end{table}

\section{Conclusion}
In this paper, we present a self-supervised framework named EMoDepth to learn monocular depth from events. Based on the observation that matching chronological events is challenging, we propose utilizing cross-model consistency from intensity frames. To fuse features more effective for improving performance, we design a multi-scale skip-connection architecture for EMoDepth. Extensive experiments on MVSEC and DSEC datasets show that EMoDepth can achieve state-of-the-art performance. In the feature work, we will explore exploiting events to remove moving objects to improve performance. Besides, a better event representation method for retaining more spatiotemporal information of events while keeping a small data size is also a promising direction.


\end{document}